\documentclass{article}

\usepackage[english]{babel}

\usepackage[letterpaper,top=2cm,bottom=2cm,left=3cm,right=3cm,marginparwidth=1.75cm]{geometry}

\usepackage{amsmath}
\usepackage{graphicx}
\usepackage{tabularx}
\usepackage[colorlinks=true, allcolors=blue]{hyperref}

\title{The order in speech disorder: a scoping review of state of the art machine learning methods for clinical speech classification}
\author{
    Birger Moëll\textsuperscript{†,1*}, 
    Fredrik Sand Aronsson\textsuperscript{†,2,3*}, 
    Per Östberg\textsuperscript{2,3}, 
    Jonas Beskow\textsuperscript{1}
}

\begin{document}
\maketitle

\noindent
\textsuperscript{1}KTH Speech, Music and Hearing \\
\textsuperscript{2}Theme Women’s Health and Allied Health Professionals, Unit of Speech and Language Pathology, Karolinska University Hospital \\
\textsuperscript{3}Division of Speech and Language Pathology, Department of Clinical Science, Intervention and Technology, Karolinska Institutet \\

\noindent
\textbf{Correspondence*:} \\
Corresponding Authors: \\
bmoell@kth.se, fredrik.sand@ki.se \\

\noindent
\textsuperscript{†}These authors have contributed equally to this work and share first authorship.

\begin{abstract}

\textbf{Background:} Speech patterns have emerged as potential diagnostic markers for conditions with varying etiologies. Machine learning (ML) presents an opportunity to harness these patterns for accurate disease diagnosis.

\textbf{Objective:} This review synthesized findings from studies exploring ML's capability in leveraging speech for the diagnosis of neurological, laryngeal and mental disorders.

\textbf{Methods:} A systematic examination of 564 articles was conducted with 91 articles included in the study, which encompassed a wide spectrum of conditions, ranging from voice pathologies to mental and neurological disorders. Methods for speech classifications were assessed based on the relevant studies and scored between 0-10 based on the reported diagnostic accuracy of their ML models.

\textbf{Results:} High diagnostic accuracies were consistently observed for laryngeal disorders, dysarthria, and changes related to speech in Parkinsons disease. These findings indicate the robust potential of speech as a diagnostic tool. Disorders like depression, schizophrenia, mild cognitive impairment and Alzheimers dementia also demonstrated high accuracies, albeit with some variability across studies. Meanwhile, disorders like OCD and autism highlighted the need for more extensive research to ascertain the relationship between speech patterns and the respective conditions.

\textbf{Conclusion:} ML models utilizing speech patterns demonstrate promising potential in diagnosing a range of mental, laryngeal, and neurological disorders. However, the efficacy varies across conditions, and further research is needed. The integration of these models into clinical practice could potentially revolutionize the evaluation and diagnosis of a number of different medical conditions.

\end{abstract}

\section{Introduction}

Speech is a cornerstone of human communication, intricately connected to our cognitive, neurological, and psychological processes. It is a complex function, involving precise coordination between the brain, periferal nervous system, and vocal organs, making it a potential window into the functioning of the human body and mind. The subtleties of speech—such as tone, pace, volume, and inflection—serve not only as conveyors of information but also as biomarkers of health. Variations in these speech characteristics can signal underlying conditions, and with the advent of machine learning (ML), we have the tools to analyze these signals with unprecedented depth and accuracy.

The process of speech production is a sophisticated one, originating from the cognitive formulation of language, which is then converted into motor plans and executed through the movement of various articulators. This process can be impacted by a plethora of factors, from structural abnormalities of the vocal organs to neurodegenerative diseases, each affecting speech in unique ways. Understanding these impacts and the nuances of speech production is vital for the advancement of diagnostic methodologies.

\subsection{Rationale and Significance}

The rationale for this review stems from the emerging evidence that speech, as a complex sensorimotor activity, can act as a biomarker for a variety of disorders. The fine-grained analysis that ML provides enables the detection of patterns that are too subtle for human perception, allowing for earlier and more accurate diagnoses. This review synthesizes current research exploring the efficacy of ML in leveraging speech for the diagnosis of disorders, highlighting both the potential and the challenges of this approach.

The significance of this work lies in its potential to revolutionize diagnostic practices. By integrating ML with speech analysis, there is an opportunity to enhance the precision of diagnoses across a spectrum of disorders, from those affecting the laryngeal system to mental and neurodegenerative conditions. This has profound implications for patient outcomes, as accurate and timely diagnoses are critical for effective treatment and management. In doing so, this review not only contributes to the scientific understanding of speech as a diagnostic marker but also underscores the need for further interdisciplinary research to refine these techniques and translate them into clinical practice.

In our review, our focus was on including a broad range of disorders from different scientific and clinical paradigms: neurological, mental and laryngeal to help create a unified perspective on speech classification using machine learning.

\subsection{Speech production}

Speech production is a complex process that integrates neural mechanisms and motor functions to facilitate verbal communication. The initiation of speech involves the organization of motor programs, which are pre-constructed codes within the brain that direct the sequences of muscular contractions required for speech \cite{guenther2016neural}. These motor programs are housed in various brain areas, with the primary motor cortex playing a pivotal role in executing voluntary motor movements, including speech \cite{hickok2012cortical}. The pre-motor and supplementary motor areas also contribute to the planning and coordination of speech movements \cite{bohland2006fmri}. Furthermore, Broca’s area, traditionally associated with expressive language production, is involved in the complex motor sequencing required for fluent speech \cite{flinker2015redefining}. These brain regions are interconnected with the basal ganglia and cerebellum, which fine-tune motor commands to produce smooth, coordinated speech \cite{ackermann2014brain}.

Once motor plans are set, speech breathing, or the control of exhalation during phonation, comes into play. It is a finely tuned process that regulates airflow and subglottal pressure, essential for generating voice \cite{ito2015neural}. The diaphragm, intercostal muscles, and abdominal muscles are actively engaged in modulating respiratory functions to accommodate the varying demands of speech \cite{baken2000clinical}. Phonation is the process by which the vocal folds produce sound and is achieved by the approximation of the vocal folds within the larynx. When air from the lungs passes between them, it causes a drop in pressure that pulls the vocal folds together, a phenomenon known as the Bernoulli effect \cite{titze2006myoelastic}. This effect, combined with the elasticity of the vocal folds, sets them into vibration. The frequency of this vibration, which determines the pitch of the sound, is primarily controlled by the length, tension, and mass of the vocal folds. In pathological conditions, such as vocal fold nodules or paralysis, the regular vibration of the vocal folds is disrupted, which can lead to a hoarse or breathy voice, demonstrating the delicate balance required for normal phonation \cite{sataloff2017voice}. Volume, or loudness, is governed by the force of the airstream provided by the lungs—the subglottal pressure—and the amplitude of the vocal fold vibration. A greater subglottal pressure causes a larger amplitude of vibration, producing a louder sound \cite{ladefoged2011course}. Additionally, the vocal fold's ability to sustain this vibration over time, known as phonatory efficiency, is essential for producing a steady and clear voice, which is a critical aspect of effective speech communication \cite{titze1994principles}.

Finally, articulation—the formation of distinct speech sounds—is achieved through the coordinated movements of the tongue, lips, jaw, and soft palate \cite{kent2000research}. These articulators adjust the shape and size of the vocal tract, which filters the sound produced by the larynx to create the consonants and vowels that compose spoken language \cite{fowler1980coarticulation}. The tongue can alter the resonance of the oral cavity and is central to vowel formation and consonant differentiation \cite{stone2016guide}, while lip movements contribute to bilabial and labiodental consonant production and influence vowel quality through rounding or spreading, also aiding in visual speech perception \cite{schwartz2014no}. Vertical movements of the jaw further modulate vowel space \cite{green2017sequential}, and the direction of the airflow can be directed by the soft palate and give rise to nasal or oral sounds \cite{feng2010role}. These articulatory actions are underpinned by auditory and somatosensory feedback mechanisms, essential for speech motor control and adaptation \cite{nasir2016somatosensory}. This feedback is important for articulatory precision and adaptability across different communicative contexts and is crucial for speech motor learning and error correction \cite{golfinopoulos2011integration}.

% \subsection{Speech production and the brain}
\begin{figure}
    \centering
    \includegraphics[width=1\textwidth]{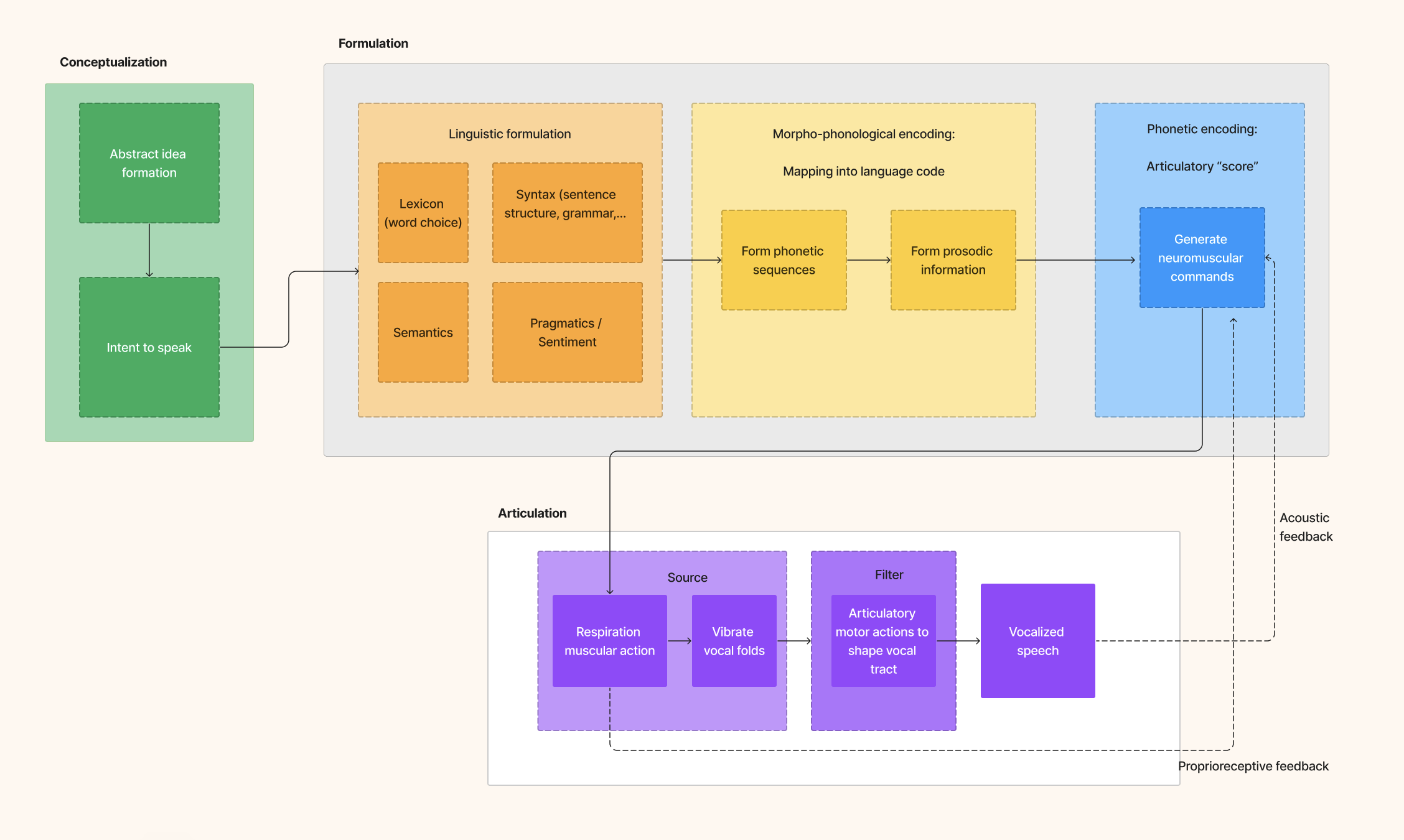}
    \caption{Schematic representation of the process of speech production adapted from \cite{voleti2019review}}
    \label{fig:speech-brain}
\end{figure}

Figure \ref{fig:speech-brain} shows the process of speech production from conceptualization to output. In this paper, our focus will be narrowed to those aspects involved exclusively in speech production. This includes morpho-phonological encoding, phonetic encoding, articulation, and acoustic feedback. We will deliberately exclude conceptualization and linguistic formulation.

Neural circuits within the brain, involved in language processing and conceptualization, mediate the process of speech production \cite{hickok2012}. Once an idea takes shape, it undergoes a translation into a linguistic representation that relies on the accurate selection of words and grammatical structures. Following this, the focus shifts to planning and programming speech movements, requiring coordination of various structures such as the vocal cords, tongue, lips, and jaw. The motor control system dictates the precise movements essential for speech production \cite{guenther2006}. The vocal tract serves as a filter for sounds emanating from the vocal cords, creating an acoustic signal recognized as speech \cite{kadiri2021}. Both auditory and somatosensory feedback loops are essential in fine-tuning speech production \cite{rochet-capellan2002}. Furthermore, memory and emotions also have a significant impact on speech production. While memory aids in word retrieval and stores vocabulary and grammatical conventions, emotions influence the prosody and lexical choices in speech \cite{ries2015, juslin2015}.

\subsection{Influence of pathology on speech production}

Various pathologies can have significant impacts on the process of speech production, due to their influence on the neural, muscular, and respiratory systems involved in the generation of speech. Parkinson's disease, for example, often leads to hypokinetic dysarthria, where speech becomes monotonous, soft, and may have a reduced rate or be rapid and stuttering, reflecting the rigidity and bradykinesia characteristic of the disease \cite{skodda2011progression}. In Alzheimer’s disease, speech rate reduction and frequent hesitations are common, with syntactic processing being relatively preserved in the early stages, indicating that phonetic disruptions occur earlier than language structure degeneration  \cite{szatloczki2015speaking}. Furthermore, the disease can result in anomia and other language deficits that affect verbal expression, with speech often becoming empty of content and riddled with paraphasias and circumlocutions as the disease progresses \cite{taler2008language}.

Schizophrenia is a chronic and severe mental disorder characterized by distortions in thinking, perception, emotions, language, sense of self, and behavior \cite{american2013diagnostic}. Individuals with schizophrenia may display disorganized speech patterns, such as derailment or incoherence, which are thought to stem from deficits in the organization and selection of thoughts \cite{covington2012schizophrenia}. This disorganization can also be mirrored in the motor aspects of speech production, resulting in erratic speech rhythm and unusual intonation patterns.

Organic changes to the vocal organs can also affect speech production negatively. Vocal nodules are small benign growths on the vocal folds leading to a raspy or hoarse voice due to irregular vocal fold vibrations \cite{silverman2019voice} \cite{sataloff2017voice}. The presence of nodules can make it difficult to maintain vocal intensity and can result in a voice that tires easily, impacting the ability to produce a sustained, clear voice \cite{verdolini2014review}.

\section{Method}
The aim of our scoping review was to identify articles that used machine learning in the classification of disorders using speech.

Our review should not be viewed as an extensive evaluation of research on a specific condition. There are several review articles for speech analysis for specific conditions such as \cite{kumari2023review} focused on Parkinson, \cite{low2020automated} focused on psychiatric disorders and \cite{cummins2015review} focused on depression and suicide risk. 

Instead our review addresses a broad range of disorders where speech is affected. Our contribution is a scoping review focused on the comprehensive examination of various disorder types, encompassing neurological, mental, and laryngeal disorders, to highlight the versatility and potential of machine learning techniques in detecting subtle and distinct speech pattern alterations associated with each disorder.

\subsection{Search Strategy}
The search was conducted using the PubMed API, leveraging a comprehensive set of search terms to identify articles relevant to the application of machine learning in diagnosing mental, laryngeal, and neurological disorders through speech analysis. The primary search strings utilized both free text, Mesh terms and combined searches. Search terms were identified in the Mesh library as well as from the indexation of published articles identified in the search process.

\subsection{Additional searches}
We included additional searches for AD and MCI focused on the Madress and Adresso AD classification challenges since AD was underrepresented in our original search.

\subsection{Inclusion Criteria}
To be included in the review, articles needed to meet the following criteria:
\begin{itemize}
    \item Published in English.
    \item Peer-reviewed.
    \item Primary focus on the application of machine learning for analyzing speech patterns.
    \item Availability of full text.
\end{itemize}

\subsection{Exclusion Criteria}
Articles were excluded if they:
\begin{itemize}
  \item Were not available in English.
  \item Did not primarily focus on speech analysis using machine learning.
  \item Review articles, opinion pieces, or case studies without original data.
    \item  Articles using transcriptions of spoken language.
\end{itemize}

We excluded all articles focused on transcription since transcription articles would shift the focus of the review to also include language.

\subsection{Data extraction using AI}
We used the Open AI API with the model gpt-3.5-turbo to extract a short summary of each article, including what speech features were used in each article and a suggestion for inclusion or exclusion.

The AI created summaries were helpful but the AI system was not capable of reliably including or excluding articles. All articles were reviewed manually and all decisions on inclusion or exclusion was done by the authors. In the future it might be possible that AI system can be capable of reviewing articles but our experiment showed that human review is still essential. 

\subsubsection{Prompt}
The following prompt was used:

\emph{
Based on the following text containing title and abstract, please write a short paragraph about why this article should be included or excluded from the systematic review regarding \textbf{{instruction}}}

\subsection{Articles found}
Our search process identified 564 potential articles,
out of which 91 were included in the final review.

Articles were excluded through a range of reasons including:

\begin{itemize}
  \item No full text found
  \item MRI, CT, EEG, Image, Wearable sensors, Video, Transcription or Multi-modal data
    \item Not focused on speech
\end{itemize}

For included articles the following information was reviewed:

\begin{itemize}
  \item Disease / Condition
  \item Method
    \item Speech features
    \item Number of participants
    \item Classification method
    \item Accuracy
    \item GRADE rating
\end{itemize}

The GRADE rating was a combination of  a subjective evaluation of the article quality including an objective evaluation of the model accuracy. If the article was deemed to be of high quality, the rating corresponded to the accuracy of the classification, with high quality papers with an accuracy over 0.9 having a grade rating of 9 and lower accuracy models or models with less clear findings receiving a lower GRADE rating.

\section{Results}
In the reviewed studies, various machine learning models were employed to analyze speech features and classify different health conditions. The studies encompassed a diverse set of diseases, including Parkinson's (PD), Mild Cognitive Impairment (MCI), Alzheimer's Disease (AD), Multiple Sclerosis (MS), spinal pain, voice diseases, Amyotrophic Lateral Sclerosis (ALS), depression, Schizophrenia, OCD, Autism and  apathy in older adults with cognitive disorders. English was the most common language but the results included studies on Chinese, Greek, Spanish, Malay and Hebrew.

\subsection{Neurological Diseases}

\begin{table}
\caption{Studies on Speech Analysis in Neurological Diseases using Machine Learning}
\label{tab:summary}
\centering
\scriptsize{ % Reducing font size
\begin{tabular}{|p{1.5cm}|p{0.5cm}|p{1.0cm}|p{3.5cm}|p{2.5cm}|p{3.0cm}|p{2.1cm}|}
\hline
\textbf{Author} & \textbf{Year} & \textbf{Disease} & \textbf{Method \& Speech Features} & \textbf{Participants} & \textbf{Classification/Method} & \textbf{Accuracy/AUC} \\
\hline
Svoboda et al. & 2022 & MS & Acoustic feature extraction using automatic phoneme segmentation & 65 MS-positive and 66 healthy & Random Forest & 0.82 \\
\hline
Rozenstoks et al. & 2019 & MS & Syllables from AMR and SMR paradigms & 120 MS patients and 60 healthy controls & CNN algorithm & 97.8\% \\
\hline
Duey et al. & 2023 & Spinal Pain & Speech recordings via smartphone, Praat features & 60 patients, 384 observations & K-nearest neighbors (KNN) & 0.71 \\
\hline
Rahman et al. & 2021 & PD & Pangram utterance, MFCC, jitter, shimmer & 464 healthy and 262 Parkinson's & Gradient-boosted decision tree & AUC 0.753 \\
\hline
Karabayir et al.& 2020 & PD & 44 speech-related acoustic features & PD patients and control subjects & Light Gradient Boosting & AUC 0.951 \\
\hline
Almaloglou et al.& 2021 & PD & Voice recordings represented as spectrograms & Individuals from mPower study & 2D CNN & 98\% \\
\hline
Gomez et al.& 2020 & PD & Glottal amplitude distributions & 48 PD patients, 48 HC, 48 NS & Hierarchical Clustering & 89.6\% \\
\hline
Lahmiri et al.& 2017 & PD & 22 voice features & 147 PD patients and 48 HC & SVM & 92\% \\
\hline
Vasquez et al.& 2018 & PD & m-FDA scale, phonation, articulation, prosody & 68 PD patients and 50 HC & i-vectors approach & correlation 69\% \\
\hline
Kuresan et al.& 2019 & PD & WPT, MFCC & PD patients & HMM and SVM & - \\
\hline
Hemmerling et al.& 2022 & PD & Vowel /a/ recordings & 27 patients & Regression & - \\
\hline
Zhang et al.& 2020 & PD & Real-time speech signal analysis & PD patients and controls & SVM and SVR & 88.74\% \\
\hline
Rahman et al.& 2021 & PD & Pangram utterance & 726 participants & XGBoost & AUC 0.753 \\
\hline
Campi et al.& 2023 & PD & EMD-GP model based on BLIMFs & PD patients & EMD-GP & >90\% \\
\hline
Doumari et al.& 2023 & PD & CSADT method & Four Parkinson’s datasets & CSADT & ~100\% \\
\hline
Moudden et al.& 2017 & PD & PCA with LR and C5.0 classifiers & LSVT Voice Rehabilitation dataset & PCA-LR and PCA-C5.0 & 100\%, 99.92\% \\
\hline
Gromicho et al.& 2022 & ALS & ALSFRS-R scores, respiratory tests & 664 ALS patients & sdtDBN & - \\
\hline
Wang et al.& 2018 & ALS & Articulatory speech features & 1831 phrases & Gradient Boosting & - \\
\hline
Stegmann et al.& 2021 & ALS & Speech acoustics via mobile app & 39 participants, 1,971 observations & Regression (FVC prediction) & 0.80 \\
\hline
Nagumo et al.& 2020 & MCI & Short-sentence reading tasks & 8779 participants & Machine learning model & AUC 0.61, 0.67, 0.77 \\
\hline
Amini et al.& 2023 & MCI & 55 speech features & 200 participants, 2800 speech samples & Machine learning model & AUC 0.02 multi class accuracy \\
\hline
Martínez et al.& 2023 & MCI & acoustic, rhytm and voice quality features &  & Machine learning model & Accuracy 0.83.3\% \\
\hline
Wang et al.& 2023 & MCI & speech acoustic features & Mandarin speakers & Random forest & AUC 0.81\% \\
\hline
Balogh et al.& 2023 & MCI & Silence and hesitations & 25 MCI 25 HC  & Receiver Operation Characteristic (ROC) analysis & - \\
\hline
Tamm et al.& 2023 & AD & eGEMAPS features & 237 participants & DNN for multilangual speech classification& 82.6\% \\
\hline
Chen et al.& 2023 & AD & openSmile paralinguistic features & 237 participants & SVM & 69.57\% \\
\hline
Jin et al.& 2023 & AD & disfluency features, wav2vec 2.0 features & 237 participants & Ensemble & 86.69\% \\
\hline
Shah et al.& 2023 & AD & word length, pause rate, speech intelligibility & 237 participants & SVM & 69\% \\
\hline
% New entries
Gauder et al.& 2021 & AD & Embeddings from trill, allosaurus, wav2vec 2.0 & 87 AD patients and 79 HC & Neural network on segmented speech & 78.9\% \\
\hline
Pan et al.& 2021 & AD & Acoustic and linguistic features from spontaneous speech & 87 AD patients and 79 HC & wav2vec 2.0, TDNN & 74.65\% (Acoustic), 84.51\% (Linguistic) \\
\hline
Agbavor et al.& 2022 & AD & Data2vec model & 87 AD patients and 79 HC & AI system for detection and severity prediction & AUC 0.846, 0.835 \\
\hline
Hason et al.& 2022 & AD & Acoustic feature analysis & 87 AD patients and 79 HC & Random Forest & 82.2\% (Prediction), 71.5\% (Stage Classification) \\
\hline
Liu et al.& 2023 & AD & Speech pause features from acoustic signals & 87 AD patients and 79 HC & Combined with traditional feature sets & Significant improvements \\
\hline
Pérez-Toro et al.& 2021 & AD & Acoustic and linguistic features & 87 AD patients and 79 HC & Early fusion strategy & Increased by 5 percentage points \\
\hline
Kashyap et al.& 2020 & Cerebellar Ataxia & Phase-based cepstral features & 42 CA patients and 23 controls & SVM & 84.6\% \\
\hline
Duey et al.& 2023 & Pain & Speech recordings via Beiwe app & 60 patients & KNN & 71\% \\
\hline
\end{tabular}
}
\end{table}

\subsubsection{Dysarthria, general}

The field of voice pathology detection and classification is witnessing rapid advancements, thanks to deep learning and artificial intelligence techniques. \cite{zaim2023accuracy} evaluated the Online Sequential Extreme Learning Machine (OSELM) algorithm \cite{huang2005line} which, when tested on the Malaysian Voice Pathology Database (MVPD), achieved remarkable accuracy in detecting dysphonic voices and differentiating vocal fold pathologies. \cite{xu2023dysarthria} devised a deep neural network (DNN) model emphasizing clinical interpretability for classifying dysarthric speakers, with the model's decisions aligning with established clinical studies. Meanwhile, \cite{song2022detection} developed an AI model, PWSI-AI-AC, to distinguish between ataxic and hypokinetic dysarthria, showcasing results that outpaced even neurology residents in classification accuracy. \cite{joshy2022automated} undertook a comprehensive comparison of deep learning techniques for categorizing dysarthria severity, with the DNN classifier using MFCC-based i-vectors producing superior results in certain scenarios. Lastly, addressing the challenges faced by current tools in analyzing short-duration speech, \cite{gupta2021residual} introduced a Residual Network (ResNet) approach, achieving an impressive 98.90\% classification accuracy on a standard dataset. Collectively, these studies underline the transformative potential of AI and deep learning in voice pathology, emphasizing their capability to provide accurate, clinically-relevant insights.

\subsubsection{Parkinson’s Disease (PD)}

Research on PD has highlighted the potential of voice recordings as a diagnostic tool, leveraging machine learning to analyze speech patterns and detect PD symptoms. Rahman et al. focused on a study where participants recited a specific pangram, achieving an AUC of 0.753 with a gradient-boosted decision tree model \cite{rahman2021detecting}. Karabayir et al. analyzed 44 speech-related acoustic features from the UCI-Machine Learning repository and achieved a remarkable AUC of 0.951 using Light Gradient Boosting \cite{karabayir2020gradient}.

Almaloglou introduced a 2D CNN method that achieved an accuracy of 98 percent in identifying PD from the mPower study's voice recordings \cite{Almaloglou2021}. Gomez et al. used Information Theory to investigate the speech alterations in PD patients, achieving an accuracy of 94.8 percent and 92.8 percent for males and females, respectively, when differentiating PD patients from healthy controls \cite{gomez2020methodology}. Lahmiri et al. assessed various machine learning techniques, with the SVM classifier demonstrating superior performance in diagnosing PD based on dysphonia measurements \cite{lahmiri2018performance}.

Vasquez et al. proposed a machine learning methodology to predict dysarthria levels in PD patients using the m-FDA scale, emphasizing the potential of the i-vectors approach \cite{vasquez2018towards}. Kuresan et al. combined MFCC and WPT (Wavelet Packet Transform) for diagnosing PD from smart device recordings, with the HMM classifier showing the best results \cite{kuresan2019fusion}. Hemmerling focused on speech signal processing to assess the neurological state of PD patients post-medication, leading to the development of therapeutic monitoring software \cite{hemmerling2022prediction}. Zhang et al. introduced a real-time speech signal tool, achieving an accuracy of 88.74 percent in diagnosing PD, integrated into the mobile app "No Pa" \cite{zhang2020intelligent}.

Rahman et al. also introduced a web-based framework for remote PD screening, with the XGBoost model performing consistently across varied environments \cite{rahman2021detecting}. Campi's study presented the EMD-GP model, achieving an accuracy above 90 percent in detecting ataxic speech disorders \cite{campi2023ataxic}. Doumari used the CSADT method for early PD diagnosis, achieving nearly 100 percent accuracy \cite{doumari2023early}. Lastly, Moudden combined PCA (Principal Component Analysis) with Linear regression and C5.0 classifiers, efficiently classifying PD patients with remarkable accuracy \cite{moudden2017feature}.

\subsubsection{Mild Cognitive Impairment} (MCI)
\cite{amini2023automated} analyzed 2800 speech recordings from 200 Greek Cypriots aged over 65, categorizing them into normal cognition, MCI, and dementia based on their Mini-mental state examination scores. From these, 55 speech features were extracted. The system achieved a multi-classification accuracy rate of an AUC of 0.92 using these speech features combined with clinical data. 

\cite{martinez2023reading} focused on several speech parameters, including acoustic features, rhythm, duration, and voice quality, to screen for MCI and Alzheimer’s disease. Using a combination of various speech parameters, the model achieved an accuracy of 83.3\%. \cite{wang2023automatic} explored the use of speech acoustic features to detect MCI in Mandarin-speaking elderly individuals. Key parameters included speech rate, utterance duration, and variability of F1 and F2. The best classification was achieved using the Random Forest classifier with an accuracy and AUC of 0.81. 

\cite{balogh_imre_gosztolya_hoffmann_pákáski_kálmán_2023} introduced an innovative approach focusing on silent segments and hesitations during verbal fluency tasks. The study found that silent parameters, such as the number and average length of silent pauses, displayed classification ability similar to traditional methods. 

\cite{nagumo2020automatic} demonstrated that as cognitive impairment intensifies, so do language impairments.

Finally, \cite{yoshii2023screening} utilized humanoid robots to collect conversations from 94 participants, extracting 17 prosodic and acoustic features. The automated classification using SVM achieved accuracies of 66.0\% and 68.1\% for MMSE and everyday conversations, respectively. Collectively, these studies highlight the potential of speech parameters in detecting cognitive decline and the promise of automation in this domain.

\subsubsection{Alzheimer's disease (AD)}
In the quest for automatic recognition of (AD) from speech, \cite{gauder2021alzheimer} extracted embeddings from models like trill \cite{shor2020towards} and wav2vec 2.0 \cite{baevski2020wav2vec}. Using a neural network on segmented speech, an impressive accuracy of 78.9\% was achieved, surpassing challenge baselines.

\cite{pan2021using} explored the use of acoustic and linguistic features from spontaneous speech for AD detection. Using ASR systems like wav2vec 2.0 and time delay neural networks (TDNN), notable accuracies of 74.65\% and 84.51\% were achieved for acoustic-only and linguistic-only models, marking improvements over prior baselines.

The AI system by \cite{agbavor2022artificial} was developed to detect AD and predict its severity. Central to this was the data2vec model \cite{baevski2022data2vec}, which when tested on internal and external datasets yielded an AUC of 0.846 and 0.835 respectively. For comparative purposes, the wav2vec 2.0 model was also employed.

Through acoustic feature analysis, \cite{hason2022spontaneous} demonstrated the potential of patient speech as an early diagnostic and monitoring tool for AD. Employing classifiers like the Random Forest, accuracies of 82.2\% for AD prediction and 71.5\% for stage classification were reported.

A unique approach by \cite{liu2023efficient} was the extraction of speech pause features from acoustic signals. Combined with traditional feature sets, significant improvements in AD classification were observed, emphasizing the importance of speech pause data.

\cite{perez2021influence} extracted both acoustic and linguistic features from AD patient speech recordings. When combined using an early fusion strategy, the accuracy increased by 5 percentage points, emphasizing the synergistic effect of combining these features.

Multilingual AD speech classification was explored by several studies. \cite{10096770} demonstrated an accuracy of 82.6\% using eGEMAPS\cite{eyben2015geneva} feature vectors. \cite{10095522} emphasized the significance of pause and speech rate, achieving an accuracy of 0.6957\%. Supporting this, \cite{10096253} combined disfluency features with wav2vec 2.0, achieving an accuracy of 86.69\%. Finally, \cite{10095593} incorporated various features, including word length and pause rate, attaining a 69\% accuracy in multilingual AD prediction.

\subsubsection{Multiple Sclerosis (MS)} 
For MS, \cite{svoboda2022assessing} identified a distinctive MS "vocal fingerprint" using the Random Forest model, emphasizing the potential of voice patterns in MS diagnosis. In another study, \cite{rozenstoks2019automated} developed a neural network algorithm to detect syllable variations in MS patients, underscoring the differences in speech patterns between MS patients and healthy controls.

\subsubsection{Amyotrophic Lateral Sclerosis (ALS)}
In ALS research, \cite{gromicho2022dynamic} used dynamic Bayesian networks to understand the impact of clinical and demographic factors on disease progression. Concurrently, \cite{wang2018automatic} predicted speaking rates in ALS patients using articulatory speech features, and \cite{stegmann2021estimation} leveraged mobile technology to assess respiratory function in ALS through speech acoustics.

\subsubsection{Cerebellar Ataxia}
\cite{kashyap2020quantitative} developed a system using phase-based cepstral features to assess speech abnormalities, differentiating ataxic speakers from controls.

\subsubsection{Somatic disorders and pain}

\cite{duey2023daily} introduced an objective pain assessment tool for neurosurgical practice, using a K-nearest neighbors machine learning model to classify pain intensity based on speech recordings.

\cite{xiao2021data} explored the potential of the formant centralization ratio (FCR) as an acoustic marker for oral tongue cancers, suggesting its utility in detecting and evaluating speech function related to these cancers.

\cite{fang2019detection} and \cite{fujimura2019discrimination} explored computerized techniques, with the former achieving 99.32\% accuracy using deep neural networks for detecting "Hot Potato Voice" - a symptom of severe upper airway obstructions. The latter utilized acoustic features like resonance characteristics and Mel-Frequency Cepstral Coefficients to identify "hot potato voice". They employed a physical articulatory speech synthesis model and a Support Vector Machine for machine learning classification, using both synthetic and real patient voice samples to diagnose upper airway obstructions with 88.3\% accuracy.

\subsection{Laryngeal disorders and voice feature detection}
\begin{table}
\caption{Studies on Speech Analysis in Laryngeal disorders using Machine Learning}
\centering
\scriptsize{ % Reducing font size
\begin{tabular}{|m{1.2cm}|m{0.8cm}|m{2.5cm}|m{3.5cm}|m{2.0cm}|m{3.0cm}|m{2.0cm}|}
\hline
\textbf{Author} & \textbf{Year} & \textbf{Disease/Condition} & \textbf{Method \& Speech Features} & \textbf{Participants} & \textbf{Classification/Method} & \textbf{Accuracy} \\
\hline
Kohler et al. & 2016 & Nodule and unilateral paralysis & Glottal signal parameters & 248 voice recordings & ANN, SVM, HMM & Peak: 97.2\% \\
\hline
Cordeiro et al. & 2017 & Physiological and neuromuscular larynx pathologies & Speech signals & Voice samples & Hierarchical classification & Peak: 98.7\% \\
\hline
Jokinen and Alku & 2017 & Spectral tilt estimation & Glottal inverse filtering (GIF) & Telephone speech & DNN & Improved accuracy \\
\hline
Reid et al. & 2022 & Pathological voices & Spectrograms & 950 training, 406 test (SVD) & VGG19 CNN + SVM & Overall: 97.8\% \\
\hline
Zakariah et al. & 2022 & Voice pathology & Chroma, mel spectrogram, MFCC & Saarbruecken Voice Database & DNN & Peak: 96.77\% \\
\hline
Lee et al. & 2022 & Vocal recovery post-surgery & Voice spectrograms & 114 patients & EfficientNet + LSTM & AUC: 0.894 (Grade) \\
\hline
Fang et al. & 2019 & Voice disorders & MFCCs & 60 normal, 402 pathological & DNN, SVM, GMM & DNN: 99.32\% (MEEI) \\
\hline
Fujimura et al. & 2019 & "Hot potato voice" & Voice spectral envelope & Patient voice samples & SVM & 88.3\% \\
\hline
Kojima et al. & 2021 & Voice quality assessment & GRBAS scale & 1,377 voice samples & 1D-CNN & Grade: 0.771 \\
\hline
Wu et al. & 2018 & Pathological vs. normal voices & Spectrograms & 482 regular, 482 dysphonic & CNN & Testing: 77.0\% \\
\hline
Suppa et al. & 2020 & Adductor-type spasmodic dysphonia & Cepstral analysis & Voice samples & Machine learning & Effective differentiation \\
\hline
Alghamdi et al. & 2022 & Voice diseases & MFCC & 687 healthy, 908 pathological & 1D and 2D-CNN & 84\% \\
\hline
Hung et al. & 2022 & Pathological voice classification & Voice features & Far Eastern Memorial Hospital datasets & SincNet system & Up to 7\% higher accuracy \\
\hline
Compton et al. & 2023 & Voice disorders & Mel-spectrograms & Voice recordings (2019-2021) & ANN & 83\% \\
\hline
Xu et al. & 2023 & Dysarthria & Four clinically relevant features & UA-Speech and TORGO databases & DNN & UA-Speech: 93.97\% \\
\hline
Song et al. & 2022 & Ataxic and hypokinetic dysarthria & Voice analysis & 804 perceptual speech analyses & CNN & High accuracy \\
\hline
Joshy et al. & 2022 & Dysarthria severity & Speech features & UA-Speech and TORGO databases & DNN, CNN & UA-Speech: 93.97\% \\
\hline
Gupta et al. & 2021 & Dysarthric speech & Short speech segments & Universal Access corpus & ResNet & 98.90\% \\
\hline
\end{tabular}
}
\end{table}

Several research endeavors have delved into using advanced computational techniques for voice pathology detection and classification. 

\subsubsection{Pathological voices }
\cite{reid2022development} used SVM on spectograms to detect voice pathology with 97.8\% accuracy while \cite{zakariah2022analytical} developed a deep neural network (DNN), using  three voice characteristics: chroma, mel spectrogram, and mel frequency cepstral coefficient (MFCC) based on audio samples of the vowels /a/, /i/, and /u/ achieving accuracy uof 96.77\%. \cite{wu2018cnn} introduced a CNN-based algorithm for screening pathological voice recordings, achieving up to 77\% accuracy. \cite{alghamdi2022neurogenerative} achieved 84\% accuracy in voice disease analysis using the Saarbruecken Voice Database.  \cite{wang2023pathological} introduced a method achieving an accuracy of 98.99\% in classifying pathological voice signals. \cite{hung2022usingsincnet} introduced an "explainable SincNet system" that outperformed traditional methods. Lastly, \cite{compton2023ai} showcased AI's superiority over otolaryngologists with an accuracy of 83\%.

\subsubsection{Nodules and vocal cord paralysis}
\cite{kohler2016analysis} utilized glottal signal parameters and achieved over 95\% accuracy in classifying voice disorders between speakers with nodule on the vocal cords;
speakers with vocal cords paralysis; and speakers with normal voices. 

\subsubsection{Voice sample identification}
\cite{cordeiro2017hierarchical} employed hierarchical classification to enhance voice sample identification, achieving an impressive 98.7\% accuracy for pathological voices. 

\subsubsection{Spectral tilt}\cite{jokinen2017estimating} innovatively estimated the spectral tilt in telephone speech using deep neural networks (DNNs), offering potential in enhancing speech intelligibility. 

\subsubsection{Vocal recovery post thyroid surgery}
\cite{lee2022predictions} predicted vocal recovery post thyroid surgery using voice spectrograms with correlations as high as 0.766. 

\subsubsection{Reinke's edemba}
\cite{naranjo2021replication} introduced replication-based regularization techniques that achieved a consistent accuracy rate of 0.89 for detecting Reinke's edema.

\subsubsection{Voice quaility assessments}
\cite{chen2023deep} employed deep learning for auditory-perceptual voice assessments with promising results while \cite{kojima2021new} combined machine learning with the GRBAS scale for voice quality assessment and \cite{fujimura2022classification} developed 1D-CNN models that paralleled human evaluations in voice quality assessment. 

\subsubsection{Spasmodic dysphonia}
\cite{suppa2020voice} showcased the efficacy of cepstral analysis and machine-learning algorithms in diagnosing spasmodic dysphonia.

\subsection{Mental disorders}

\begin{table}
\caption{Studies on Speech Analysis in Mental Disorders using Machine Learning}
\centering
\label{tab:psychiatric_disorders}
\scriptsize{ % Reducing font size
\begin{tabular}{|m{1.8cm}|m{0.5cm}|m{2.05cm}|m{2.5cm}|m{2.0cm}|m{3.5cm}|m{1.5cm}|}
\hline
\textbf{Author} & \textbf{Year} & \textbf{Disease/Condition} & \textbf{Method \& Speech Features} & \textbf{Participants} & \textbf{Classification/Method} & \textbf{Accuracy} \\
\hline
Dougdu et al. & 2022 & Emotion Recognition & openSMILE feature sets & Berlin Database & SVM with SMO, MLP, LOG & 87.85\% \\
\hline
Mocanu et al. & 2021 & Emotion Recognition & Spectrogram inputs & RAVDESS, CREMA-D & SE-ResNet with GhostVLAD & 83.35\% \\
\hline
Zhang et al. & 2023 & Emotion Recognition & Gender-specific features & 2660 speech samples & CNN, BiLSTM & 87.91\% \\
\hline
Wasserzug et al. & 2023 & Depression & Over 200 raw voice parameters & 144 participants & Behavioral vocal analysis & Significant \\
\hline
Du et al. & 2023 & Depression & LPC and MFCC & Two public datasets & MSCDR & 0.86 \\
\hline
Tonn et al. & 2022 & Depression & Non-content speech parameters & 163 participants & VoiceSense tool & 0.41 (correlation) \\
\hline
Lu et al. & 2021 & Depression & Residual thinking with attention & SRE paradigm & Attention-based residual network & High \\
\hline
Othmani et al. & 2022 & Depression & Audiovisual data & DAIC-Woz dataset & Model of Normality &  87.4\% \\
\hline
Zhao et al. & 2022 & Depression & Multimodal markers & AVEC 2017, AVEC 2019 & Cross-modal attention mechanism & - \\
\hline
Zhang et al. & 2022 & Depression & Multimodal machine learning & Real-world data & Coarse-grained and fine-grained models & High \\
\hline
Kim et al. & 2023 & Depression & Korean speech features & 318 participants & CNN & 78.14\% \\
\hline
Tonn et al. & 2022 & Depression & Verbal communication nuances & 163 participants & VoiceSense tool & 0.41 (correlation) \\
\hline
Teixeira et al. & 2023 & Schizophrenia & Emotional state through speech emotion & Narrative review & - & - \\
\hline
Huang et al. & 2023 & Schizophrenia & Acoustic features & 130 individuals & Classification algorithms in scikit-learn & 0.815 \\
\hline
Tan et al. & 2021 & Schizophrenia & Speech aberrations & 89 participants & Machine learning & High \\
\hline
De et al. & 2023 & Schizophrenia & Acoustic speech parameters & 284 participants & Machine-learning algorithm & 86.2\% \\
\hline
Fu et al. & 2021 & Schizophrenia & Sch-net with CBAM & 56 participants & Deep learning & 97.68\% \\
\hline
Clemmensen et al. & 2022 & OCD & Vocal activation & 64 participants & ANOVA, logistic regression & 80\% \\
\hline
Tang et al. & 2023 & Delirium & Acoustic and textual elements & Older adults & Machine learning & 78\% \\
\hline
Rybner et al. & 2022 & Autism & ML models across contexts & - & Evaluation & - \\
\hline
Xu et al. & 2022 & Multiple Disorders & Speech, facial expressions, body movements & 228 participants & Machine-learning methods & 84.7\% \\
\hline
Sumali et al. & 2020 & Dementia vs. Depression & Acoustic features & - & SVM with LASSO & High \\
\hline
Konig et al. & 2019 & Apathy in Older Adults & Narrative speech tasks & Older adults & Correlation analysis & 0.88 (Males) \\
\hline
\end{tabular}
}
\end{table}

\subsubsection{Depression}
In diagnosing depression, a range of studies are leveraging vocal characteristics to detect and gauge severity. \cite{wasserzug2023development} employed behavioral vocal analysis, demonstrating a clear distinction in vocal patterns between Major Depressive Disorder patients and controls. Similarly, \cite{du2023depression} and \cite{tonn2022digital} introduced models emphasizing comprehensive representation and non-content speech parameters, both showing promise in the accurate detection of depressive states. Adding to this, \cite{lu2021speech} introduced an attention-based residual network that outperforms traditional methods by using speech features related to negative questions under negative emotions, showing high accuracy for both male and female subjects. \cite{othmani2022model} and \cite{zhao2022unaligned} used audiovisual and multimodal data, respectively, to predict depression and its relapse with promising results. \cite{zhang2022adolescent} enhanced the accuracy of adolescent depression detection through multimodal machine learning. Moreover, \cite{kim2023automatic} demonstrated that speech data from Korean speakers can be analyzed to predict depression with a high degree of accuracy using a deep convolutional neural network model.

\subsubsection{Schizophrenia}
Schizophrenia detection is also benefiting from voice analysis. While \cite{teixeira2023narrative} explored EEG features and speech parameters, \cite{huang2023} and \cite{tan2021investigating} leveraged acoustic features to diagnose schizophrenia with notable accuracy. \cite{de2023acoustic} and \cite{fu2021sch} further underscored the potential of automated speech analysis for diagnosis and differentiation.

\subsubsection{Emotion recognition}\cite{dougdu2022comparison} highlighted the efficacy of machine learning algorithms for Vocal Emotion Recognition, with Support Vector Machines and Neural Networks emerging as top performers. Meanwhile, \cite{mocanu2021utterance} and \cite{zhang2023deep} introduced innovative models that enhance emotion recognition through robust feature representation and gender-specific emotion recognition, respectively.

\subsubsection{Comparison between disorders}
The potential to distinguish between psychiatric disorders using voice analysis was highlighted by \cite{xu2022identifying} and \cite{sumali2020speech}. Both studies showcased the capability of speech characteristics in differentiating disorders like schizophrenia and depression.

\subsubsection{Other mental disorders}
In other psychiatric domains, \cite{clemmensen2022associations} and \cite{tang2023characterizing} demonstrated the predictive power of vocal features in diagnosing conditions like OCD and delirium. The study by \cite{rybner2022vocal} emphasized the need for caution in generalizing machine learning models across diverse contexts, particularly in autism research. Lastly, the study on older adults with cognitive disorders by \cite{konig2019detecting} demonstrated the correlation between speech characteristics and apathy, emphasizing the role of voice analysis in detecting cognitive disorders.

\begin{table}
\caption{Main findings with GRADE system analysis}
\centering
\footnotesize{ % Reducing font size
\renewcommand{\arraystretch}{1.2} % Slightly increase row height for readability
\begin{tabularx}{\textwidth}{|p{3.5cm}|p{1.5cm}|p{1.5cm}|p{2.5cm}|X|}
\hline
\hline
\textbf{Disease/Condition} & \textbf{No. of Articles} & \textbf{Score (0-10)} & \textbf{GRADE Analysis} & \textbf{Explanation} \\
\hline
Nodule and unilateral paralysis & 1 & 9 & High & High accuracy reported, suggesting speech can be a reliable marker. \\
Physiological and neuromuscular larynx pathologies & 1 & 9 & High & Extremely high accuracy suggests strong potential for speech analysis. \\
Voice pathology (general) & 3 & 8 & Moderate & High accuracy across studies, but results may vary depending on the specific pathology. \\
Voice disorders & 2 & 8 & Moderate & High accuracy but may vary depending on the specific disorder. \\
Dysarthria & 3 & 9 & High & High accuracy across studies, making speech a reliable marker for diagnosis. \\
Emotion Recognition & 3 & 7 & Moderate & Accuracy is good, but emotional states can be transient and influenced by many factors. \\
Depression & 9 & 8 & Moderate & Consistent research showing high accuracy, but results can be affected by the individual's current emotional state. \\
Schizophrenia & 6 & 8 & Moderate & High accuracy in some studies, but individual results can vary. \\
OCD & 1 & 6 & Low & Limited research, and the relationship between OCD and speech patterns isn't firmly established. \\
Delirium & 1 & 7 & Moderate & Good accuracy, but more research is needed for a definitive conclusion. \\
Autism & 1 & 6 & Low & Limited research available on using speech patterns for autism diagnosis. \\
Apathy in Older Adults & 1 & 7 & Moderate & High correlation, but the specific relationship between apathy and speech needs more research. \\
Dementia vs. Depression & 1 & 7 & Moderate & High accuracy in differentiating, but still requires more comprehensive studies. \\
Multiple Sclerosis (MS) & 1 & 8 & Moderate & High accuracy suggests speech can be a potential diagnostic marker. \\
Parkinson’s & 10 & 9 & High & A lot of research with consistent high accuracy makes speech a promising diagnostic tool. \\
ALS & 2 & 8 & Moderate & Good accuracy, but more research is needed to understand the specific speech patterns. \\
Mild Cognitive Impairment & 5 & 7 & Moderate & Promising research in the last years. \\
Alzheimers Dementia & 11 & 7 & Moderate & Good relative accuracy but usually improved when combined with language based classification \\
Cerebellar Ataxia & 1 & 8 & Moderate & Good accuracy suggests potential in using speech for diagnosis. \\
Pain in Neurological Disease & 1 & 7 & Moderate & Good accuracy, but more comprehensive studies are needed. \\
\hline
\end{tabularx}
}
\end{table}

\subsection{Main findings}
These findings demonstrate the significant potential of machine learning models in analyzing speech features for classification in a diverse range of etiologies. The studies achieved varying degrees of accuracy, with some reaching over 90\%, indicating the promise of these approaches in contributing to healthcare diagnostics. Parkinson's and voice disorders where the conditions with the greatest accuracy reported by machine learning models.

\section{Discussion}

Our review shows that speech can be a powerful tool for classification in a wide range of disorders. Regarding what speech features to focus on, here as some trends we found in the articles we reviewed.
\subsection{How you say it}
For disorders where changes in speech was prominent such as laryngeal disorders, PD and depression, many successful algorithms used acoustic measurements from the voice source. Short tests such as phonating a single vowel could be useful for classification in several disorders.

\subsection{Silence is golden}
In disorders where speech changes were more subtle such as MCI and AD, successful algorithms focused on finding speech features related to cognitive functioning such as speech rate and pause rate. Notably, several articles focused on silence features, such as amount of pauses and pause length that could be useful for classification even in multi-lingual settings. Pause length can be seen as a useful feature for understanding the underlying cognitive system. Pauses gives information about tempo which correlates to cognitive load. Speech rate also holds this information but might be harder to untangle from what is being said. Several different methods have been used in previous research. \cite{hoffmann2010temporal} found that hesitation pauses significantly differed between control groups and mild AD cases. \cite{sluis2020automated} noted increased pause lengths correlating with more severe cognitive symptoms. \cite{yuan2021pauses} successfully used pause structure and language models to classify 89.6\% of participants accurately.

\subsection{Cognitive Load and Brain Entropy}

Research in (MCI) and dementia suggests that non-linguistic speech features, such as pauses, can reflect changes in brain function. Cognitive load, which measures the mental effort used in cognitive tasks, is closely linked to the concept of brain entropy \cite{CarhartHarris2018EntropicBrain}—the variability in neuronal activity. Changes in speech, like increased pauses and slower tempo, can indicate a high cognitive load and a potential shift in brain entropy levels.

Simpler speech patterns and longer response times may reveal cognitive strain and hint at the brain's capacity to manage entropy, i.e. the brain's ability to process information in a timely manner. Seemingly simple measurements of pauses and speech rate might hint at the underlying workings of the brain through cognitive load which itself is an indication of brain entropy. The confusion often observed in patients with AD can be viewed as a manifestation of disrupted entropy processing. This disruption can lead to a decline in the brain's ability to maintain optimal information processing, which in turn may be reflected in the observed speech patterns.

The balance between cognitive load and brain entropy is delicate, and in the context of neurodegenerative diseases such as AD, this balance is often compromised. As the brain's entropy handling becomes less efficient, patients exhibit more pronounced changes in speech. These changes are not just symptomatic of the disease, but also serve as markers for the altered information processing capacity of the brain.

In essence, the relationship between cognitive load, brain entropy, and speech provides a framework for voice analysis as a diagnostic tool for neurological conditions. By analyzing speech patterns and their correlation with brain entropy, researchers and clinicians may be able to detect early signs of cognitive impairment. The continued exploration of these links could lead to more refined speech-based diagnostic methods for cognitive disorders, potentially offering a non-invasive and cost-effective means of early detection and monitoring the progression of these conditions.

\subsection{Practical aspects of integrating ML models in clinical practice}

\subsubsection{Collection of clinical speech data}
The collection of clinical speech data for developing speech-based clinical models faces significant challenges, particularly concerning data quality and subsequent model performance.

Privacy concerns in clinical settings often lead to limited data capture, resulting in datasets that are not only scarce but also lack diversity and depth. This limitation is crucial because high-quality, varied data is essential for training robust models. The restricted data often fails to encompass the wide range of linguistic nuances found in clinical speech, including varying accents, terminologies, and speech patterns affected by medical conditions.

The direct consequence of compromised data quality is the diminished reliability and accuracy of speech-based clinical models. Models trained on limited or poor-quality data struggle to generalize effectively in real-world clinical environments. They may fail to accurately recognize or interpret critical speech nuances, which can significantly impact their practical utility in healthcare settings.

\subsubsection{The Imperative of Explainability in Healthcare Applications}

In clinical environments, the interpretability of machine learning models is not merely advantageous but essential. Explainability amplifies the likelihood of successful integration of AI systems into healthcare by enhancing the effectiveness of human-in-the-loop approaches, which are often indispensable for such applications.

With the increasing adoption of deep learning, we anticipate a surge in the use of (DNNs) for both feature extraction and classification tasks within the domain. However, the explainability of DNNs poses a significant challenge due to their extensive parameterization, which obscures the interpretative process. It is crucial to establish mechanisms that ensure these models remain comprehensible, enabling clinicians to grasp the reasoning behind specific predictions made by the AI. This transparency is key to maintaining trust in and ensuring the ethical application of AI technologies in sensitive fields such as healthcare.

\subsubsection{Incorporating ML based speech assessment in clinical practice}
A successful integration of ML based speech assessment is a multidisciplinary effort that requires experts in machine learning, speech assessment and clinical diagnostics.

For successful integration of ML clinicians need to overcome irrational fear of job displacement from AI and ML practitioners need to focus attention on building tools that help clinicians since healthcare needs human in the loop AI systems to remain safe in clinical settings.

Human in the loop systems where AI tools automatically assess speech files recorded in a clinical setting is a good first step. As systems mature and models become more capable, speech based systems can be used for initial assessments in a cost-effective manner. 

Since ML tools rely on data it is essential that a data pipeline is used where clinical data can be used to train models. This faces several ethical and regulatory challenges such as patient consent and GDPR. Using synthetic data can be a promising way forward that sidesteps these issues by using synthetic speech data with artificial speech impairments created to be relevant for each disorder. 

\subsection{Ecological validity}
Most articles reviewed had their own setup for data collection in a clinical setting. However some articles, notably articles related to AD where from the Madress \cite{Luz2023Multilingual} and Adresso \cite{Luz2021Detecting} challenge used similar datasets. Challenges are a great way to rally resources behind a cause but are by definition removed from clinical practice.

\subsection{Limitations}
This study is subject to several limitations that warrant consideration. Firstly, the scope of our literature review, despite being extensive, was not exhaustive. Our search process identified 564 potential articles, out of which 91 were included in the final review. It is conceivable that additional relevant studies exist that were not captured within our search parameters. While we endeavored to construct a comprehensive overview that accurately reflects the state of the art, we acknowledge the possibility of omissions.

Secondly, our study deliberately concentrated on the analysis of speech parameters to the exclusion of language content. The complex interplay between speech and language is well-documented, yet our approach necessitated a separation of these elements. This delineation was adopted to isolate and examine the specific relationship between speech characteristics and various disorders. It is important to note that this distinction, while beneficial for the focused scope of our investigation, does not fully encapsulate the intricate interactions inherent in speech-language phenomena. Future research could benefit from integrating both speech and language variables to provide a more holistic understanding of the communicative aspects related to health conditions.

\subsection{Challenges of machine learning for speech based diagnostics}

\subsubsection{Gender differences in speech}
Gender differences in voice and speech patterns can significantly affect the accuracy of machine learning models used for classifying speech. Several studies showed different performance for male and female speakers but most studies did not discuss this at all. Ensuring that training datasets are representative of all genders and employing validation techniques to check for bias are essential steps in developing fair and effective diagnostic tools. Without careful consideration of these factors, models may perform inconsistently across genders, potentially affecting diagnostic outcomes.

\subsubsection{Overfitting}
One potential bias in the studies we reviewed could be relate to overfitting, where the model learns its training data too well. For instance if training and test data is not separated well, the model will have high accuracy during evaluation but poor performance on unseen data.

\subsubsection{False negatives}
For speech based diagnostics false negatives are important to consider. Speech can be healthy while the individual might suffer from a serious health condition. As such speech based diagnostics needs to be an addition to other diagnosis methods and should not be implemented as a stand alone solution.

\subsection{Future research}

\subsubsection{Current Research Gaps}
One of the main research gaps we found was the narrow focus on a single disorder during classification. In the medical computer vision domain, models can reliably handle many, if not all disorders that are prevalent. If speech diagnostics aims to become established in  a similar way, we as researchers needs to move from single classification to multi-class classification of disorders.

Another research gap is the lack of clinical implementations. Even though research is promising use of machine learning in clinical practice is rare. Researchers need to focus attention on bringing research findings to the clinic and implement it in day to day practice.

A disorder that is a good candidate for increased focus is ALS. The impact on motor neurons guarantees speech impairment as the disorder progresses and speech could potentially be used to quickly assess the severity of the disorder.

Autism and ADHD are other disorders where more research is needed. We only included a single study on autism and no studies on ADHD but potentially speech in both disorders could be linked to neurological functioning. Prosody changes in autism and speech changes in ADHD can be seen as speech based biomarkers of the disorders. These can be used for both diagnostics and speech based training to improve functioning, by shaping prosody in autism and lowering speech rate and improving speech intelligibility in ADHD.

\subsubsection{Fusion of Multiple Modalities} 
Integrating voice data with other biological markers and health data types could significantly improve the precision of diagnoses. By creating multimodal diagnostic frameworks, clinicians may gain a more holistic view of a patient's health status.

\subsubsection{Real-time Analysis} 
Future advancements may enable real-time processing of voice samples, offering immediate feedback and the possibility of on-the-spot diagnostics. This could revolutionize response times in clinical settings, allowing for faster treatment decisions.

\subsubsection{Ethical AI and Privacy Considerations} 
As voice analysis technologies advance, it becomes increasingly critical to address data security and ethical considerations in model development. Ensuring the privacy of patient data and the ethical use of AI will be central to maintaining trust in these technologies.

\subsubsection{Personalized Diagnostic Models} 
Developing models that adapt to individual voice patterns could significantly enhance diagnostic accuracy. Personalized medicine approaches that incorporate voice analytics can lead to more tailored and effective treatment plans.

\subsubsection{Integration with Wearable Technology} 
Coupling voice analysis capabilities with wearable technology could provide continuous, non-intrusive health monitoring. This integration has the potential to alert individuals and healthcare providers to health changes in real-time.

\subsubsection{Cross-linguistic and Cultural Studies} 
Expanding research to encompass a diverse range of languages and dialects will improve the generalizability and applicability of voice diagnostic models across different populations, reducing bias and improving equity in healthcare.

\subsubsection{Advances in Interpretable AI} 
Transparency in AI decision-making processes remains a priority, especially in healthcare settings. Efforts to improve the interpretability of AI models will be essential to ensure clinicians and patients can understand and trust AI-supported diagnostic decisions.

\subsubsection{Utilization of Synthetic Data in Clinical Research}
Synthetic data, generated through algorithms that simulate real patient data, offers a solution to privacy concerns and data scarcity. It enables researchers to access diverse, high-quality datasets without compromising patient confidentiality. In speech research voice cloning, creating a voice from a single voice sample and fine-tuning, creating a voice from a base voice model and a voice dataset are techniques that can be used to create synthetic voice data. This innovation could accelerate the development of more robust and accurate AI models in voice diagnostics, allowing for extensive testing and validation in varied clinical scenarios without the ethical and legal constraints associated with real patient data. Furthermore, for rare conditions with few cases, synthetic data presents an opportunity to create models based on a more limited dataset.

\subsection{Novel insights}
Our review show that several techniques for speech classification are effective for many different disorders and we believe that it is now up to the scientific community to focus attention on implementing these findings in clinical practice.

One surprising finding is that the decision on what machine learning methods to use seemed less important for the results. Most papers implemented standard machine learning models that can be implemented in python in a few lines of code. Random Forest, K-nearest neighbour, Support Vector Machines and Gradient boosted decision trees are all easy to implement with great results based on our review.

Similarly the method for recording speech and the speech features used were also fairly straightforward. Many papers used MFCCs, Mel Spectograms or Opensmile features on spontaneous speech with good results.

This suggests that speech based clinical machine learning is indeed within reach for researchers with access to a recording device and sufficient programming skills.

\subsection{Conclusion}
This review article illustrates how machine learning models might be utilized in deciphering speech characteristics for diagnostic purposes across a spectrum of disorders. With certain studies reporting accuracy rates surpassing 90\%, the promise of these methodologies in augmenting healthcare diagnostics is indisputable. Notably, conditions such as Parkinson's disease and various voice disorders have witnessed the highest accuracy rates, underscoring the effectiveness of machine learning in these areas. However, a large number of the included studies only used a binary classification task. This can mirror some applications in health care such as a screening procedure, but more research in multiclass classification is needed - especially for related or similar conditions.

Speech is an invaluable biomarker for classifying a wide array of health conditions. The trends observed in our review indicate that acoustic measurements are critical in diseases where speech alterations are pronounced, whereas features related to cognitive functioning, such as speech rate and pause patterns, are more telling in conditions with subtler speech changes.

The implementation of speech classification in clinical settings will require a concerted effort from the scientific community to translate these findings into practice. Tools like OpenSMILE have proven effective in feature extraction, and machine learning algorithms available in Scikit-learn are robust for forming baseline models. However, the move towards multi-class classification of disorders remains a pivotal challenge to be addressed. Furthermore, the necessity for explainability in healthcare applications cannot be overstated, especially as deep learning models gain traction. It is imperative that these models remain interpretable to ensure clinicians can rely on AI-assisted decisions.

\bibliographystyle{plain}  % Eller "unsrt", "alpha", "apalike", "ieeetr" beroende på stil
\bibliography{references}

%%% Make sure to upload the bib file along with the tex file and PDF
%%% Please see the test.bib file for some examples of references

\appendix

\section{Machine learning for speech diagnostics}

Machine learning has emerged as a powerful tool in the realm of health diagnostics, especially in the domain of speech analysis. The intricate patterns and nuances in speech, which may be imperceptible to the human ear, can be detected and analyzed by machine learning algorithms, offering insights into the health of an individual, particularly their cognitive and neurological status.

\subsection{Processing of the Speech Signal}
Speech signals, captured continuously, undergo processing to render them suitable for analysis, with common techniques being detailed below.

\subsection{Mel-Frequency Cepstral Coefficient (MFCC)}
MFCCs are derived through a sequence of transformations on the original speech signal. 

1. \textbf{Framing:} The time-discrete speech signal \( s(n) \) is divided into overlapping frames of length $N$ and multiplied with a window function  \( w(n) \) where $0<n<N$:
\[ s_{frame}(n) = s(n) \cdot w(t - lT) \]
where \( l \) is the frame number and \( T \) is the frame step (also called hop length).

2. \textbf{Fourier Transform:} Each frame undergoes the Short Time Fourier Transform (STFT) to obtain its frequency spectrum \( S(m) \):
%\[ S(f) = \int_{-\infty}^{\infty} s_{frame}(t) e^{-j2\pi ft} dt \]

\[S(m) = \sum_{n=0}^{N-1}s_{frame}(n)e^{-\frac{j2\pi nm}{N}}\]

3. \textbf{Mel Filter Bank:} The spectrum is multiplied by a set of Mel filters \( M_k(m) \) to get \( Y_k \):
\[ Y_k = \sum{f} |S(m)|^2 M_k(m) \]
where \( k \) denotes the k-th Mel filter.

4. \textbf{DCT:} The log energies from the Mel filters are then processed using the Discrete Cosine Transform (DCT) to get the MFCCs:
\[ c_i = \sum_{k=1}^{K} \log(Y_k) \cos\left( \frac{\pi i (k - 0.5)}{K} \right) \]
where \( K \) is the number of Mel filters, and \( i \) is the coefficient index.

\subsection{Mel Spectrogram}
Mel spectrograms represent the short-time power spectrum of sound in the Mel scale. They serve as an intermediary step in computing the MFCCs but can also be used directly as a feature for various speech and audio processing tasks. 
For Mel Spectrogram calculation, steps 1 through 3 are identical to MFCC computation (see above), followed by  

4. \textbf{Log Compression:} The Mel power spectrum is then compressed using a logarithmic scale to obtain the Mel spectrogram:
\[ S_{mel}(k) = \log(Y_k + \epsilon) \]
where \( \epsilon \) is a small constant to prevent the logarithm from blowing up at zero.

The resulting \( S_{mel} \) provides a time-frequency representation of the speech signal that is perceptually weighted to approximate the human ear's response.

The integration of machine learning in speech analysis holds immense potential for health diagnostics, especially in the realm of cognitive and neurological health. As research in this domain continues to evolve, it is anticipated that machine learning will play an increasingly pivotal role in shaping the future of health diagnostics through speech analysis.

In this paper we will explore State of the Art in speech diagnostics using machine learning through a search of the available scientific literature.

\subsection{Machine learning models}

This section provides an overview of several machine learning models, encompassing both traditional algorithms such as Random Forest and Support Vector Machine, as well as deep learning architectures like Convolutional Neural Networks and Deep Neural Networks. 

\textbf{Random Forest (RF):} 
An ensemble learning method constructing multiple decision trees during training. For a given input vector \( x \) and a set of decision trees \( \{ T_1, T_2, ..., T_n \} \), the RF prediction, \( Y \), is given by:
\[ Y(x) = \frac{1}{n} \sum_{i=1}^{n} T_i(x) \]

\textbf{K-nearest neighbors (KNN):} 
A non-parametric algorithm where an object's classification is determined by the class most common among its \( k \)-nearest neighbors in a dataset \( D \).

\textbf{Gradient-boosted decision tree \& XGBoost:} 
An ensemble technique where new models predict the residuals of prior models. 
%Given a loss function \( L(y, F(x)) \), where \( y \) is the true label and \( F(x) \) is the prediction, the model is:
%\[ F(x) = F(x) + \eta \cdot \arg\min_{h} \sum_{i=1}^{n} L(y_i, F(x_i) + h(x_i)) \]
%where \( \eta \) is the learning rate.

\textbf{Support Vector Machine (SVM):} 
A model used for classification or regression by finding the hyperplane that best divides the dataset into classes. For a training set \( \{(x_1, y_1), ..., (x_n, y_n)\} \), SVM solves:
\[ \max_{w, b, \xi} \frac{1}{\|w\|} - \sum_{i=1}^{n} \xi_i \]
subject to \( y_i(w \cdot x_i + b) \geq 1 - \xi_i \) for all \( 1 \leq i \leq n \).

\textbf{Linear Regression:} 
A method modeling the relationship between a dependent variable and independent variables. For a dataset \( \{(x_1, y_1), ..., (x_n, y_n)\} \), it estimates the coefficients in:
\[ y = b_0 + b_1 x_1 + b_2 x_2 + ... + b_p x_p + \epsilon \]
where \( \epsilon \) is the error term.

\subsection{Deep Neural Network (DNN)}

A Deep Neural Network (DNN) is a multi-layered structure of artificial neurons, where each neuron computes a weighted sum of its inputs, passes it through an activation function, and sends the output to the next layer.

1. \textbf{Weighted Sum:} For a given neuron \(i\) in layer \(l\), the weighted sum \(z^l_i\) is given by:
\[ z^l_i = \sum_j w^l_{ij} a^{l-1}_j + b^l_i \]
where \(w^l_{ij}\) is the weight from neuron \(j\) in layer \(l-1\) to neuron \(i\) in layer \(l\), \(a^{l-1}_j\) is the activation of neuron \(j\) in layer \(l-1\), and \(b^l_i\) is the bias for neuron \(i\) in layer \(l\).

2. \textbf{Activation Function:} The activation \(a^l_i\) of neuron \(i\) in layer \(l\) is given by:
\[ a^l_i = f(z^l_i) \]
where \(f\) is a non-linear activation function, such as the sigmoid, ReLU, or tanh.

3. \textbf{Network Output:} For a DNN with \(L\) layers, the output \(y\) is the activation of the neurons in the last layer:
\[ y = a^L \]

4. \textbf{Loss Function:} To train a DNN, a loss function \( \mathcal{L}(y, \hat{y}) \) is used to measure the difference between the predicted output \(y\) and the true label \( \hat{y} \). Common loss functions include mean squared error for regression tasks and cross-entropy for classification tasks.

5. \textbf{Backpropagation:} This is the process used to update the network's weights and biases based on the computed loss. The gradient of the loss function with respect to each weight and bias is computed using the chain rule, and the weights and biases are updated in the direction that minimizes the loss.

The DNN learns the optimal weights and biases during the training phase by iteratively updating them using an optimization algorithm, typically some variant of gradient descent, to minimize the loss function.

\section{Methods used in the articles}
Our hopes with this review is to help researcher implement solutions using speech classification in a clinical setting. As such we aim to briefly go through speech features and algorithms that were effective for clinical speech classification from the articles we reviewed.

\subsection{Feature extraction}
Open smile \cite{10.1145/1873951.1874246} is a software library for automatically extracting features for use in classification. In many of the review articles, features extracted from open smile was useful for classification. Open Smile is available in python through the open smile python package.

Apart from open smile, many articles had hand crafted features. Hand crafting features is a great way to leverage knowledge in the domain to improve performance as well as keeping your features understandable.

Several articles relied on feature embeddings extracted from pretrained neural networks such as wav2vec 2.0 and data2vec. Even though embeddings show good performance they often lack explainability which might make them problematic for use in a clinical setting.

\subsection{Machine learning models}

Scikit-learn \cite{scikit-learn}  is a python library containing many machine learing algorithms including random forest, support vector machines and gradient boosted deicision trees, the most common machine learning models used in our review.

Combining open smile features and gradient boosted decision trees, random forest or support vector machines is a good contender for a baseline model when working with speech classification.

Speech datasets are often relative small which makes these models perform better compared to deep learning based models.

\end{document}